\pdfoutput=1
\documentclass[conference]{IEEEtran}
\IEEEoverridecommandlockouts
\usepackage{amsmath,amssymb,amsfonts}
\usepackage{graphics}
\usepackage{xcolor}
\usepackage{textcomp}
\usepackage{multirow}
\usepackage{xcolor}
\usepackage{mathtools}
\usepackage{inputenc}
\usepackage{caption}
\usepackage{xcolor}
\usepackage{hyperref}

\def\BibTeX{{\rm B\kern-.05em{\sc i\kern-.025em b}\kern-.08em
    T\kern-.1667em\lower.7ex\hbox{E}\kern-.125emX}}
\DeclareMathOperator{\E}{\mathbb{E}}

\def\footnoterule{\relax
	\kern-5pt
	\hbox to \columnwidth{\hfill\vrule width 0.8\columnwidth height 0.4pt\hfill}
	\kern4.6pt}
\makeatother

\title{On the Self-Similarity\\
	 of Natural Stochastic Textures
}

\author{\IEEEauthorblockN{Samah Khawaled\IEEEauthorrefmark{1} and
		Yehoshua Y. Zeevi\IEEEauthorrefmark{2}}}
\begin{document}
\maketitle

\begin{abstract}
Self-similarity is the essence of fractal images and, as such, characterizes natural
stochastic textures. This paper is concerned with the property of self-similarity in the statistical sense in the case of fully-textured images that contain both stochastic texture and structural (mostly deterministic) information. We firstly
decompose a textured image into two layers corresponding to its texture and structure, and show that the layer representing the stochastic texture is characterized by random phase of uniform distribution, unlike the phase of the structured information which is coherent. The uniform distribution of the the random phase is verified by using a suitable hypothesis testing framework. We proceed by proposing two approaches to assessment of self-similarity. The first is based on patch-wise calculation of the mutual information, while the second measures the mutual information that exists across scales. Quantifying the extent of self-similarity by means of mutual information is of paramount importance in the analysis of natural stochastic textures that are encountered in medical imaging, geology, agriculture and in computer vision algorithms that are designed for application on fully-textures images.
\end{abstract}


\section{Introduction}
 Natural images that are dominated by edge-contour structural information exhibit non-Gausssian distribution and are characterized by high kurtosis. This fact can be verified by analysis of the 1D marginal histograms of the wavelets coefficients \cite{b1,b2}. In contrast, images that incorporate natural stochastic textures (NST) exhibit Gaussian behavior, and they are characterized by the statistical properties of non-stationary and self-similarity \cite{b3}.\\
\indent The 2D Fractional Brownian Motion (fBm), introduced by Mandelbrot and Van Ness in 1968 \cite{b4}, a self-similar non-stationary Gaussian process, has been shown to be a suitable model for many NST images\cite{b5,b6}. Indeed, NST-images are endowed with the two main characteristics of fBm, namely self-similarity, in a statistical sense and Gaussianity. Self-Similarity, or scale invariance, implies existence of exact or approximate, resemblance or reoccurrence of detailed structures of features across scales. In the case of NST images, they exhibit self-similarity in the statistical sense, which is a typical characteristic of fractals \cite{b4}. This means that the image across scales shows the same statistical properties and appears to look alike.\\
\indent In contrast to structural natural images, the spatial phase of pure NST image is random and it has negligible effect compared with that of the magnitude on the appearance of the image. However, in the case of images that incorporate both NST and structural information, i.e. edges and contours, the spatial phase is of a significant importance, much more than the magnitude \cite{b18}.\\
\indent In this paper, we consider fully textured images that incorporate both pure NST and structural information. The latter is represented in the spatial frequency domain by the local phase \cite{b18}. Firstly, the images are decomposed into their structural and NST layers. We then show that the NST layer has Gaussian behavior and its spatial phase is of uniform distribution. this is tested using a suitable hypothesis framework. The focus of this paper is on NST self-similarity, in a statistical sense and on its assessment using the metric of mutual information \cite{b19}. Self-similarity of natural textures was previously discussed in various applications like textures classification and synthesis evaluation \cite{b7}, where the structural information was utilized for its assessment. Here we introduce two different approaches to quantify the degree of self-similarity. In the first approach, we address \textit{spatial} and \textit{resolution} based measurement of mutual information, whereby the former quantifies the similarity between non-overlapped patches and the latter measures it between images scales. We also show that for pure NST images the spatial similarity is preserved along scales, i.e. scale invariance is satisfied according to this criterion. In the second approach, we use the mutual information extracted from the gray scale co-occurrence matrix (GLCM)\cite{b8} for measuring the self-similarity.\\
\indent In our analysis, we images from the small subset of Kylberg texture database are used \cite{b10}. The dataset includes 240 fully-textured images belonging to 6 different substances.

\section{Background}
\subsection{Fractional Brownian Motion}\label{Fbm}
\label{sec:FBM}
\indent Fractional Brownian Motion (FBm), $B_{\text{H}}(t)$ is a continuous-time
Gaussian process characterized by the following covariance: 
\begin{equation}
{\E}\left[B_{\text{H}}(t)B_{\text{H}}(s)\right]=\frac{\sigma_{\text{H}}^{2}}{2}(|t|^{2\text{H}}+|s|^{2\text{H}}-|t-s|^{2\text{H}}),\label{eq:1}
\end{equation}
where 
\begin{equation}
\sigma_{\text{H}}^{2}=\frac{\sigma_{w}^{2}}{2}\frac{\cos(\pi\text{H})}{\pi\text{H}}\Gamma(1-2\text{H}),\label{eq:2}
\end{equation}
$\sigma_{w}^{2}$ is a known variance. The Hurst parameter $\text{H}\in(0,1)$, also
called the Hurst exponent \cite{b4}, determines the smoothness of the motion (texture in our case); higher values lead to a smoother texture. The first order increment
of the process, $G_{\text{H}}(t)=B_{\text{H}}(t+1)-B_{\text{H}}(t)$, is known as the fractional Gaussian noise (fGn). Since the fBm process is non-stationary, it is easier to study its increments, the fGn, which are stationary. The stationarity of the fGn lends itself to simple analysis and synthesis of images.\\
\indent The 2D generalization of the fBm process, called Lévy fractional Brownian field, $B_{\text{H}}(X)$, is statistically isotropic\cite{b11}. Its auto-correlation is : 
\begin{equation}
{E}\left[B_{\text{H}}(X)B_{\text{H}}(Y)\right]=\frac{\sigma_{\text{H}}^{2}}{2}(||X||^{2\text{H}}+||Y||^{2\text{H}}-||X-Y||^{2\text{H}})\label{eq:3}
\end{equation}
where $X$ and $Y$ are two points belonging to the corresponding 2D Euclidean space,
$X=(x_{1},x_{2})$ and $Y=(y_{1},y_{2})$, $||\cdotp||$ is the Euclidean
norm operator and $\sigma_{\text{H}}^{2}$ is defined in \eqref{eq:2}. 
\subsection{Self-Similarity}\label{ss}
Self-Similarity, or scale invariance in the context of fractals, means existence of exact or approximate resemblance between consecutive scales in the statistical sense\cite{b4}, as it expressed by: 
\begin{equation}
B_{\text{H}}(\alpha\boldsymbol{x})=|\alpha|^{\text{H}}B_{\text{H}}(\boldsymbol{x})\label{eq:4}
\end{equation}
where the equality is in the statistical sense. In the context of images, self-similar image preserves the same statistics i.e. PDF and moments along scales. This property implies that the image at various scales depicts the same statistical properties.
\section{Separation of Texture from Structure}\label{sec:sep}
\begin{figure}[!t]
	\begin{minipage}[b]{0.31\linewidth}
		\centering
		\centerline{\includegraphics[width=3.6cm,height=3.5cm]{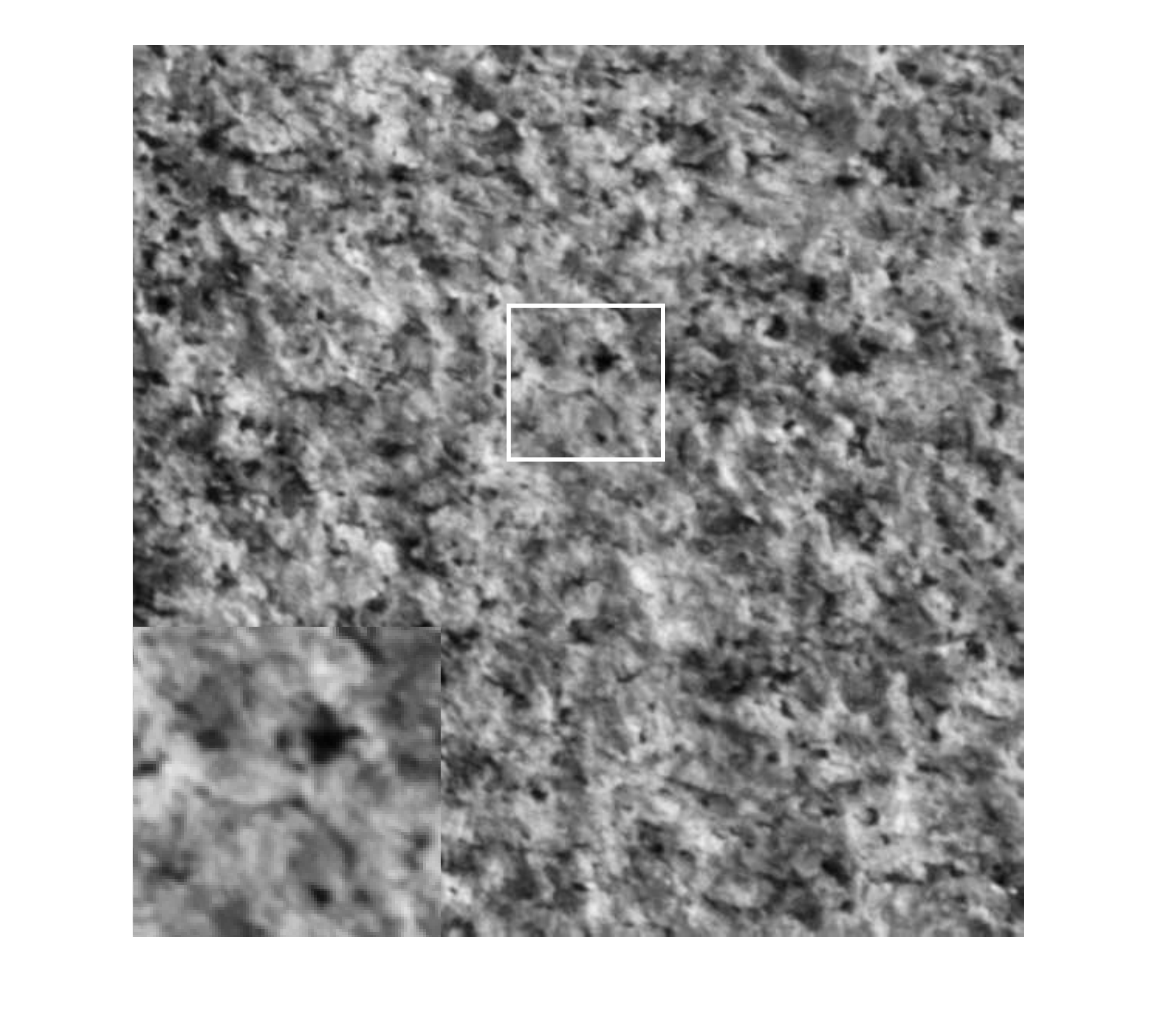}}
		\centerline{(a) \footnotesize }\medskip
	\end{minipage}
	\hspace{0.01mm}
	\begin{minipage}[b]{0.31\linewidth}
		\centering
		\centerline{\includegraphics[width=3.6cm,height=3.5cm]{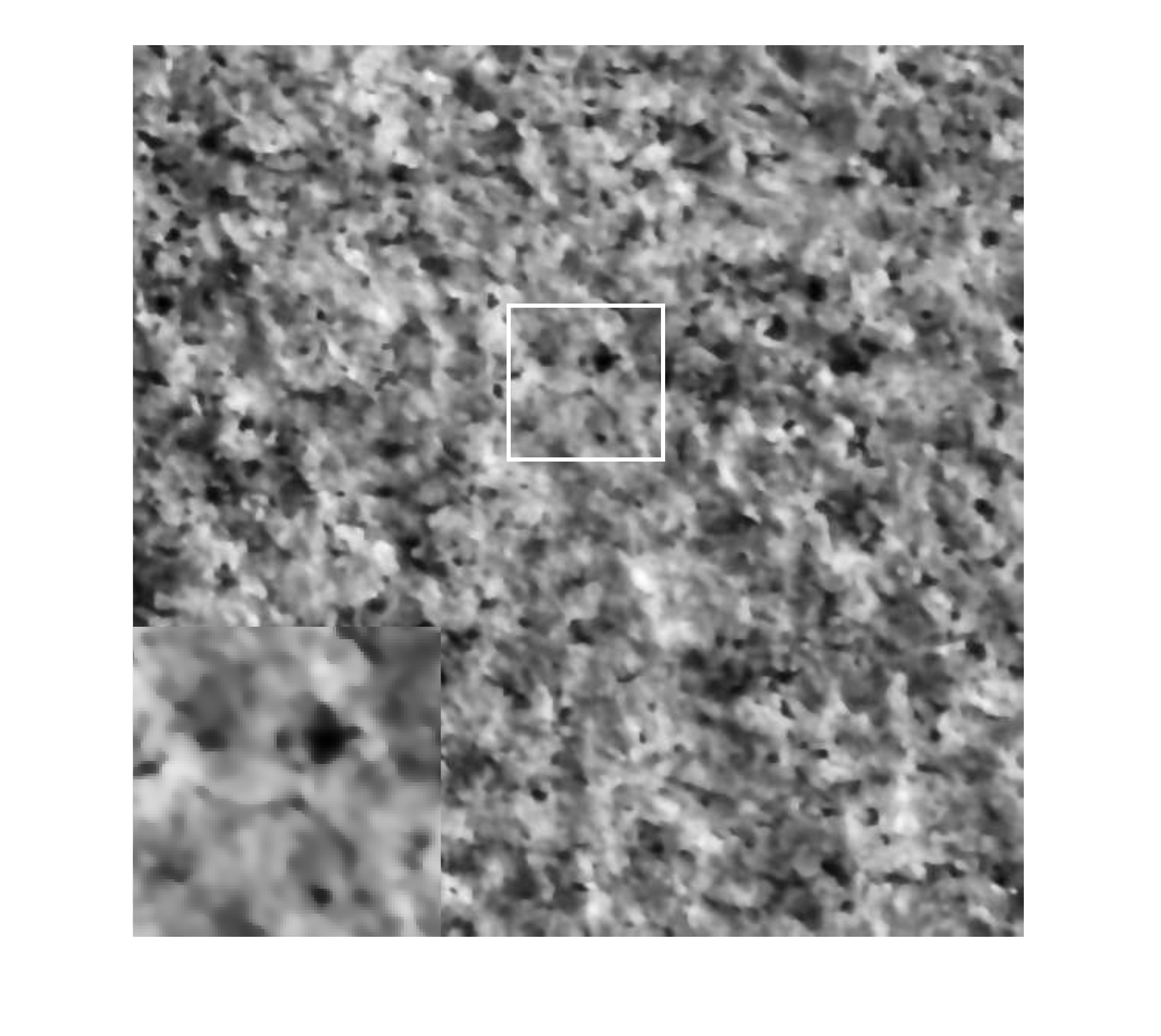}}
		\centerline{(b) \footnotesize}\medskip
	\end{minipage}
	\hspace{0.01mm}
	\begin{minipage}[b]{0.31\linewidth}
		\centering
		\centerline{\includegraphics[width=3.6cm,height=3.5cm]{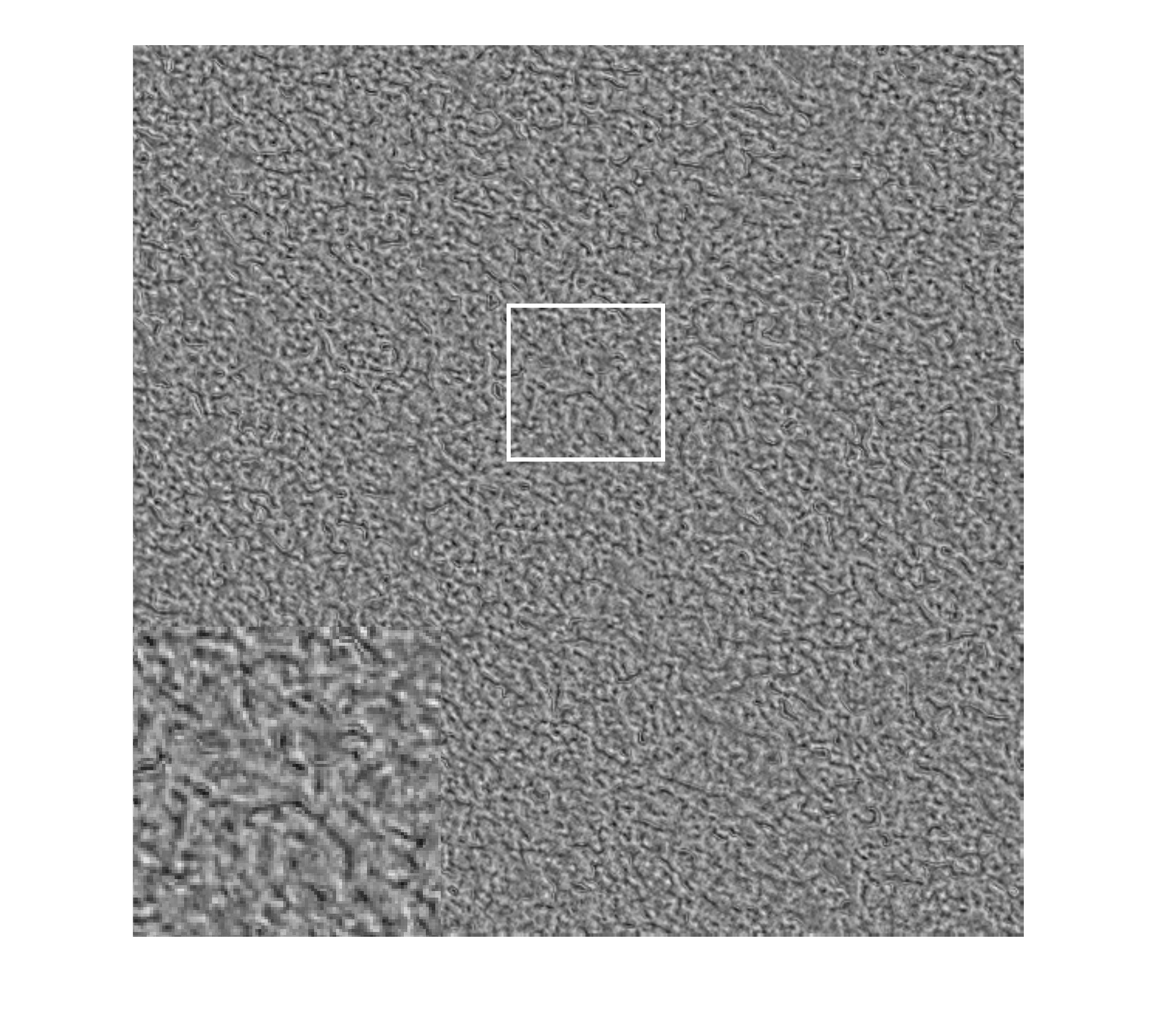}}
		\centerline{(c) \footnotesize }\medskip
	\end{minipage}
	\caption{Example of texture-structure separation. (a) Fully-textured image of stone. (b) and (c) The corresponding separated structure (deterministic) and textural (stochastic) layers, respectively. The structural layer contains only smooth areas and edge-like details, whereas the (NST) layer carries the textural information.}
	\label{fig:Sep}
\end{figure}

\indent Non-linear diffusion, such as that of Perona-Malik (PM) \cite{b9}, or the forward-and-backward (FAB) diffusion \cite{b20}, is widely used in filtering fine details and noise, while persevering or even enhancing contours and edges. We use such non-linear diffusion in our decomposition of the image into its stochastic texture layer and structured image content. Let $I$ be the input image and $\mathcal{D}_{N}$, denote the action of PM-type diffusion operated with predetermined  number of iterations, $N$. Then, the non-linearly diffused image, $I^{S}=\mathcal{D}_{N}I$ contains the structural information, while the residual image $I^{T}=I-\mathcal{D}_{N}I$ represents the texture layer. An example of this decomposition is illustrated in Fig.\ref{fig:Sep}(a-c).

\section{Statistics of NST}
\begin{figure}[!t]
	\begin{minipage}[b]{0.41\linewidth}
		\centering
		\centerline{\includegraphics[width=4cm]{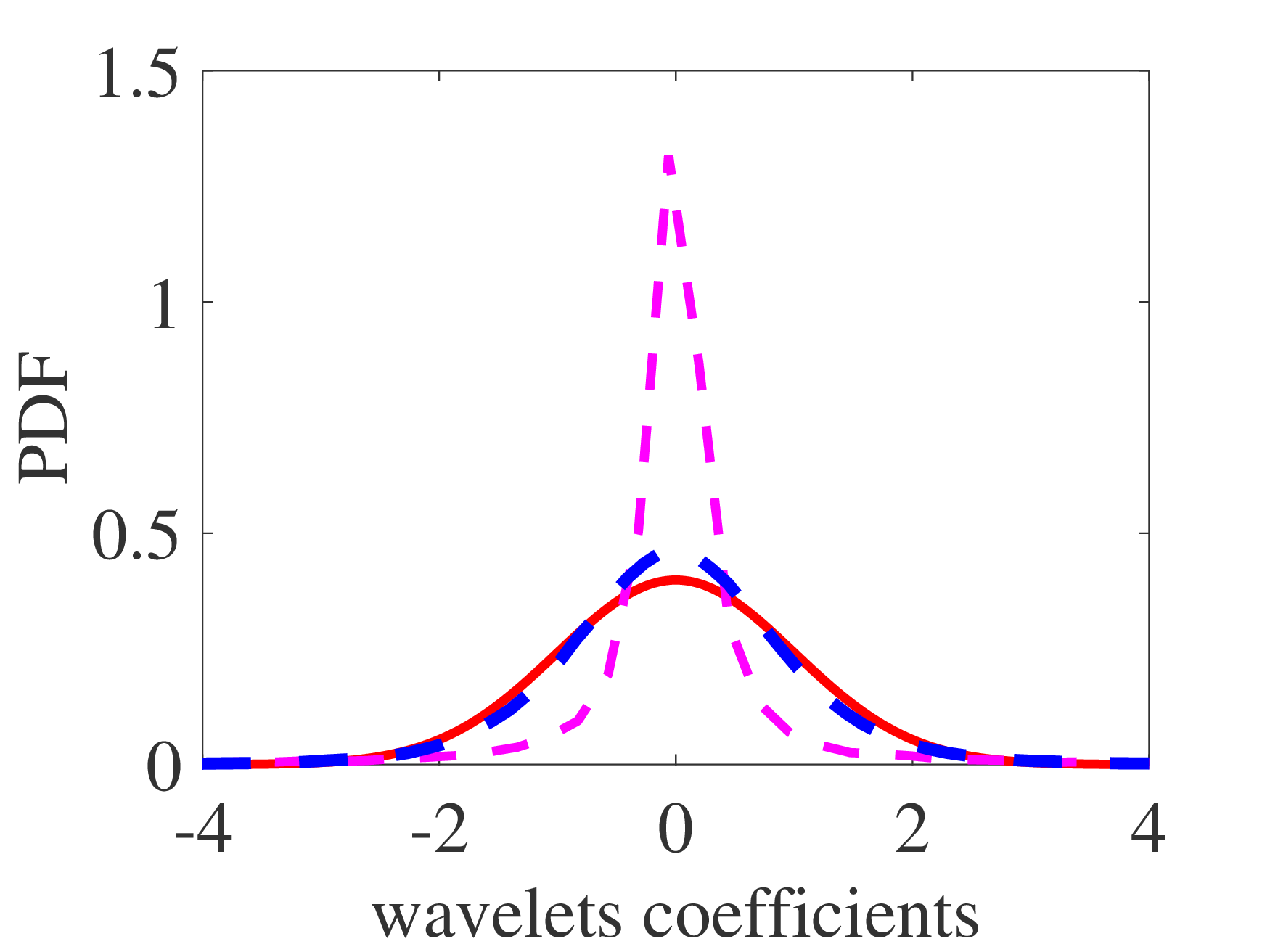}}
		\centerline{(a) \footnotesize PDF}\medskip
	\end{minipage}\label{fig:1a}
	\hspace{0.1mm}
	\begin{minipage}[b]{0.41\linewidth}
		\centering
		\centerline{\includegraphics[width=4cm]{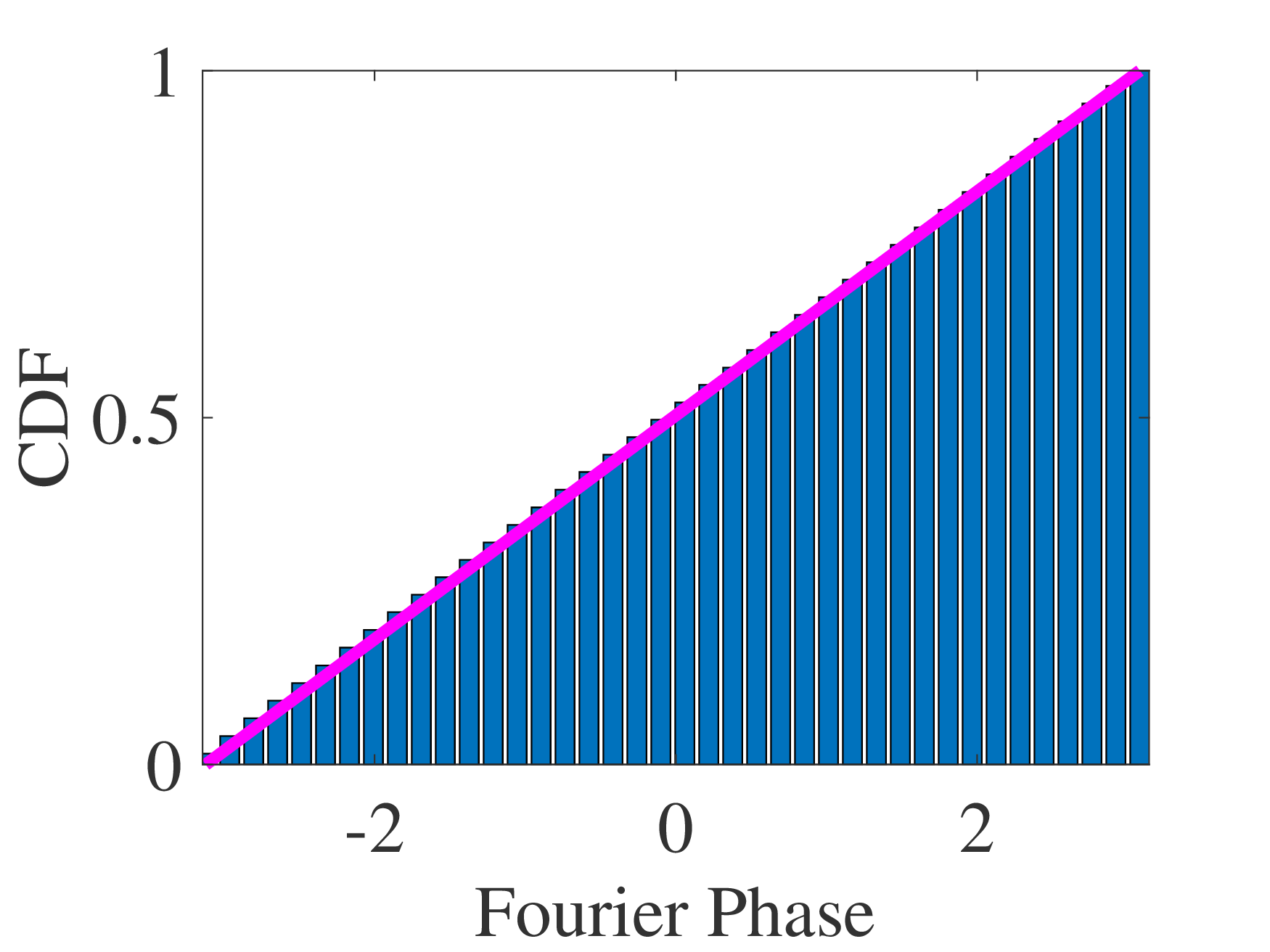}}
		\centerline{(b) \footnotesize CDF }\medskip
	\end{minipage}\label{fig:1b}
	\\
	\begin{minipage}[b]{0.41\linewidth}
		\centering
		\centerline{\includegraphics[width=4.2cm,height=4cm]{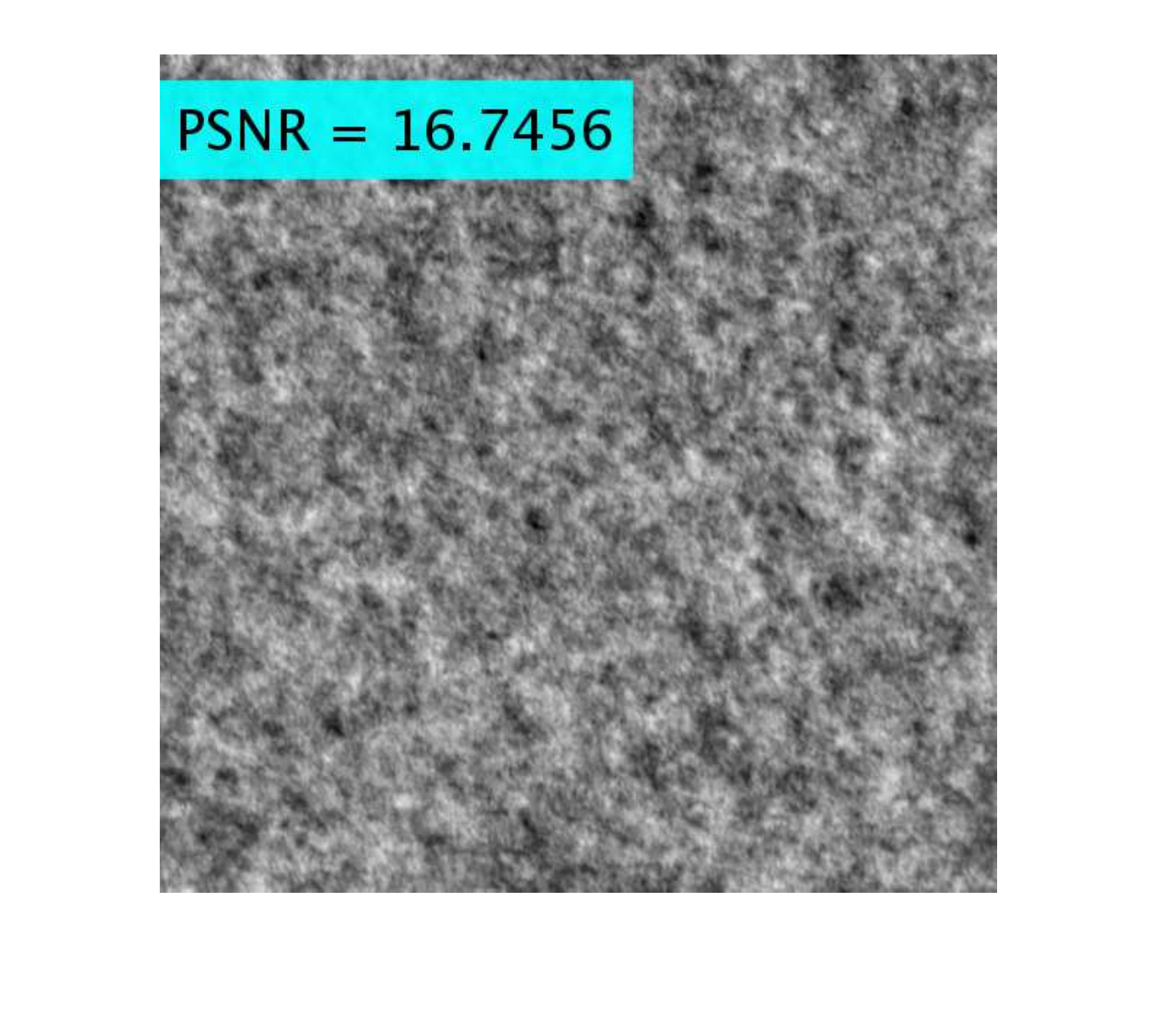}}
		\centerline{(c) \footnotesize }\medskip
	\end{minipage}
	\hspace{0.1mm}
	\begin{minipage}[b]{0.41\linewidth}
		\centering
		\centerline{\includegraphics[width=4.2cm,height=4cm]{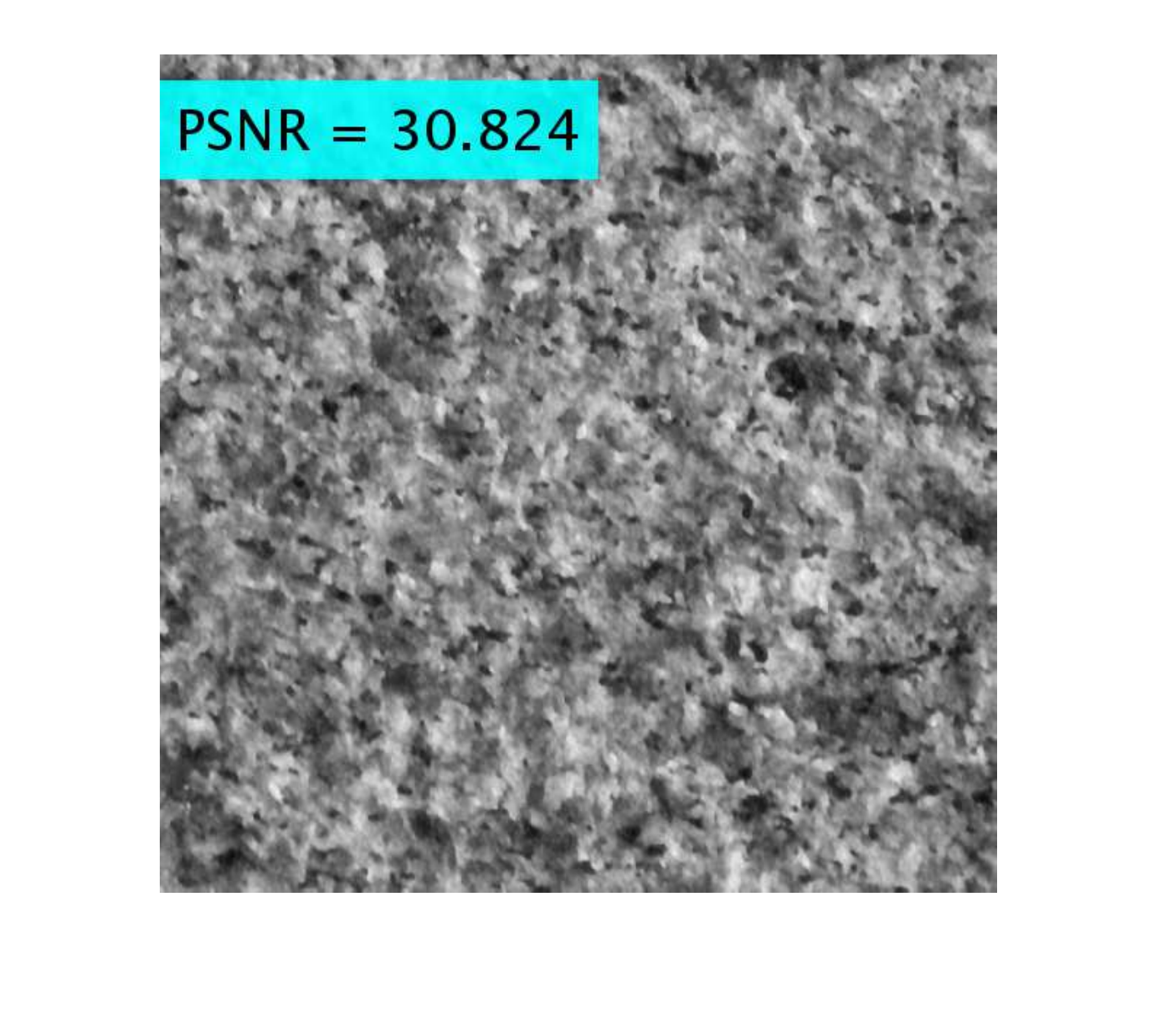}}		
		\centerline{(d) \footnotesize  }\medskip
	\end{minipage}
	\caption{Testing Gaussianity, uniformity of phase distribution and the effect of phase randomization on the structured and textured image components: (a) PDF of wavelets coefficients of the textured layer (dashed blue) with $K=4.9$ and the structured layer (dashed pink) with $K=29.4$, plotted with the best Gaussian fit PDF (solid red). (b) Cumulative distribution function (CDF) of the spatial phase, in Fourier domain , of the NST layer. Empirical CDF fits the linear line representing the best uniform-fit. (c) and (d): The result of phase randomization of the stochastic layer and the structured one, respectively}
	\label{fig:statistics}
\end{figure}

\indent The separated NST is considered to be a realization of 2D fBm process. Unlike structured natural images, it is characterized by Gaussianity and its spatial phase is random. For the purpose of our analysis, we focus primary on these two properties.   
As far as Gaussianity, the wavelets coefficients of the textural layer show Gaussian behavior. (Fig.\ref{fig:statistics}(a), dashed blue line). In contrast to NST, the wavelets' coefficients of the  structural  layer, carrying edge-like information, is characterized by high kurtosis (denoted by $K$) or heavy-tailed PDF. (See Fig.\ref{fig:statistics}(a), where the empirical PDF highlighted in dashed pink line).

\begin{table}[!b]
	\caption{Gaussian and Uniform Distribution Testing results}\label{table:1}
	\begin{centering}
			\scalebox{1.4}{
		\begin{tabular}{|c|c|c|}
			\hline 
			\multirow{2}{*}{\textbf{G}} & Entire image & $23.7\ensuremath{\%}$
			\tabularnewline
			\cline{2-3} 
			& Textural layer & $86.25\ensuremath{\%}$
			\tabularnewline
			\hline 
			\textbf{U} & Textural Layer & $92.5\ensuremath{\%}$
			\tabularnewline
			\hline 
		\end{tabular}}
		\par
	\end{centering}
\end{table}
 
\indent To substantiate the Gaussian assumption, we propose the following hypothesis testing framework to formally test if the wavelet coefficients of the NST are of Gaussian distribution. Under the null hypothesis, $\mathit{\mathsf{H_{0}}}$, it has Gaussian distribution, whereas under the alternative hypothesis, $\mathit{\mathsf{H_{1}}}$, it is of any other distribution. To this end, we use the Kolmogorov–Smirnov (KS) test \cite{b12}; one of the more widely-used tests. It compares the empirical cumulative distribution function (cdf) of the given samples, with that of a uniform distribution. It confirms $\mathit{\mathsf{H_{0}}}$ when the supremum of the difference between the two cdfs is smaller than a predetermined bound. (We used  KS-test with significance level of 0.05). \\
\indent To investigate the importance of the layer separation on Gaussianity, we performed KS-test twice. First we tested the Gaussianity of the Haar wavelet coefficients of the NST layer extracted from Kylberg images\cite{b10}. We then tested the Gaussianity of Haar wavelet coefficients of the unseparated image. The results of the tests are summarized in Table.\ref{table:1}. Whereas only $23.7\ensuremath{\%}$ of the images passed the Gaussianity test when the original (unseparated) image was used, over $86\ensuremath{\%}$ of the images passed it when only the separated texture was considered. As expected, separating texture from structure is essential for observing the Gaussian behavior of the almost-pure NST.\\  
\indent The fact that the structure is well removed from the almost-pure NST by the diffusion-based technique, is verified by having only marginal effect of the spatial phase on the image appearance. We further show that the spatial phase of the textural layer is uniformly distributed over the interval $\left[-\pi,\pi\right]$. This is confirmed empirically by observing the cumulative density function (CDF), (Fig.\ref{fig:statistics}(b)). Similarity to the Gaussianity testing, we verify the uniform distribution of the NST spatial phase by performing the KS test with significance level of 0.05. The uniform distribution hypothesis was valid for 222 out of 240 images, i.e. $92.5\%$ of the dataset shows uniform distributions.\\      
\indent We analyze the importance of phase on the appearance of NST and structured images by randomizing the phase of the structured and textured components and estimating its effect on the extent of distortion on the images. Randomizing the phase of the textured component does not yield an observable distortion in the resultant image in contrast with the profound effect on the structured one, (see Fig.\ref{fig:statistics} (c-d)).

\begin{figure}[!t]

	\begin{minipage}[c][4cm][b]{0.45\linewidth}
		\scalebox{0.95}[1.1]{
			\begin{tabular}{cccc}
				\hline 
				{\small{}$\boldsymbol{n}$} & \textbf{\small{}$\text{\textbf{FBm}}$} & \textbf{\small{}$\text{\textbf{non-NST}}$} & \textbf{\small{}$\text{\textbf{NST}}$}\tabularnewline
				\hline 
				\hline 
				\textbf{\small{}4} & {\small{}4.15} & {\small{}3.1} & {\small{}4.22}\tabularnewline
				\hline 
				\textbf{\small{}3} & {\small{}6.09} & {\small{}4.42} & {\small{}6.22}\tabularnewline
				\hline 
				\textbf{\small{}2} & {\small{}7.54} & {\small{}5.54} & {\small{}7.64}\tabularnewline
				\hline 
				\textbf{\small{}1} & {\small{}7.49} & {\small{}6.2} & {\small{}7.48}\tabularnewline
				\hline\\
		\end{tabular}}
		\centerline{(a) \footnotesize }
	\end{minipage}
	\hspace{0.1mm}
	\begin{minipage}[c][4cm][b]{0.6\linewidth}
		\centering
		\centerline{\includegraphics[width=4.4cm,height=3cm]{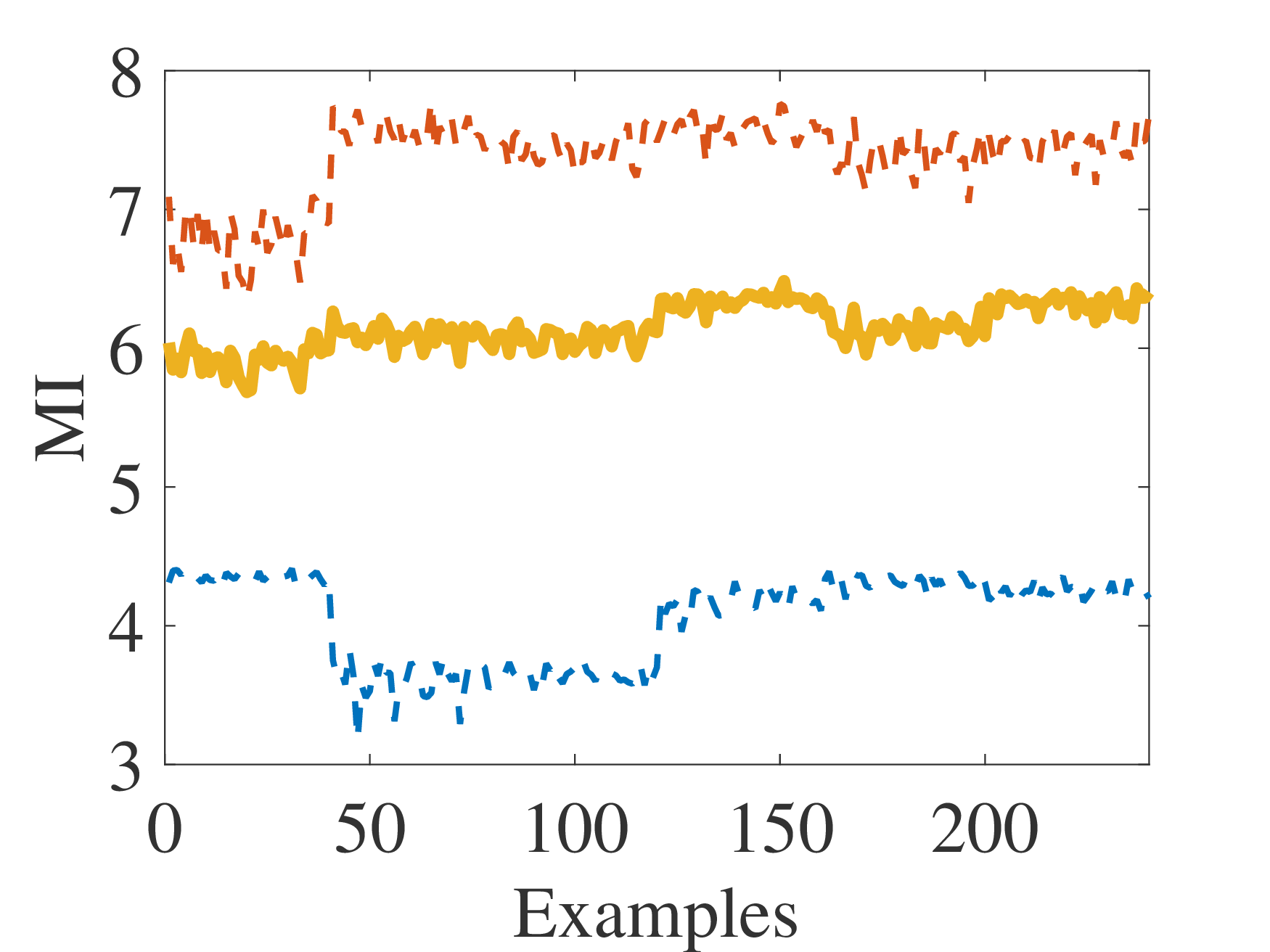}}	
		\centerline{(b) \footnotesize }
	\end{minipage}
	\caption{Scale-wise MI measurements results. (a) Calculated MI along levels for synthetic \textbf{fBm} with $\text{H}=0.2$, pure \textbf{NST} (presented in Fig\ref{fig:Sep}(d)) and \textbf{non-NST}, obtained by extraction the textural layer of \textit{cameraman} image. (b) Average, maximal and minimal MI along levels, extracted from the entire database. (Highlighted in solid-orange, dashed-red and dashed-blue, respectively). Note that the presented examples organized by classes and their MI values show a difference between the corresponding sub-classes.}
	\label{fig:MIscales}
\end{figure}
\section{Assessment of Self Similarity}
\subsection{Mutual Information-based Approach}
\begin{figure}[!t]
\begin{minipage}[b][2.5cm][t]{0.3\linewidth}
	\centering
	\centerline{\includegraphics[width=3.1cm,height=1.9cm]{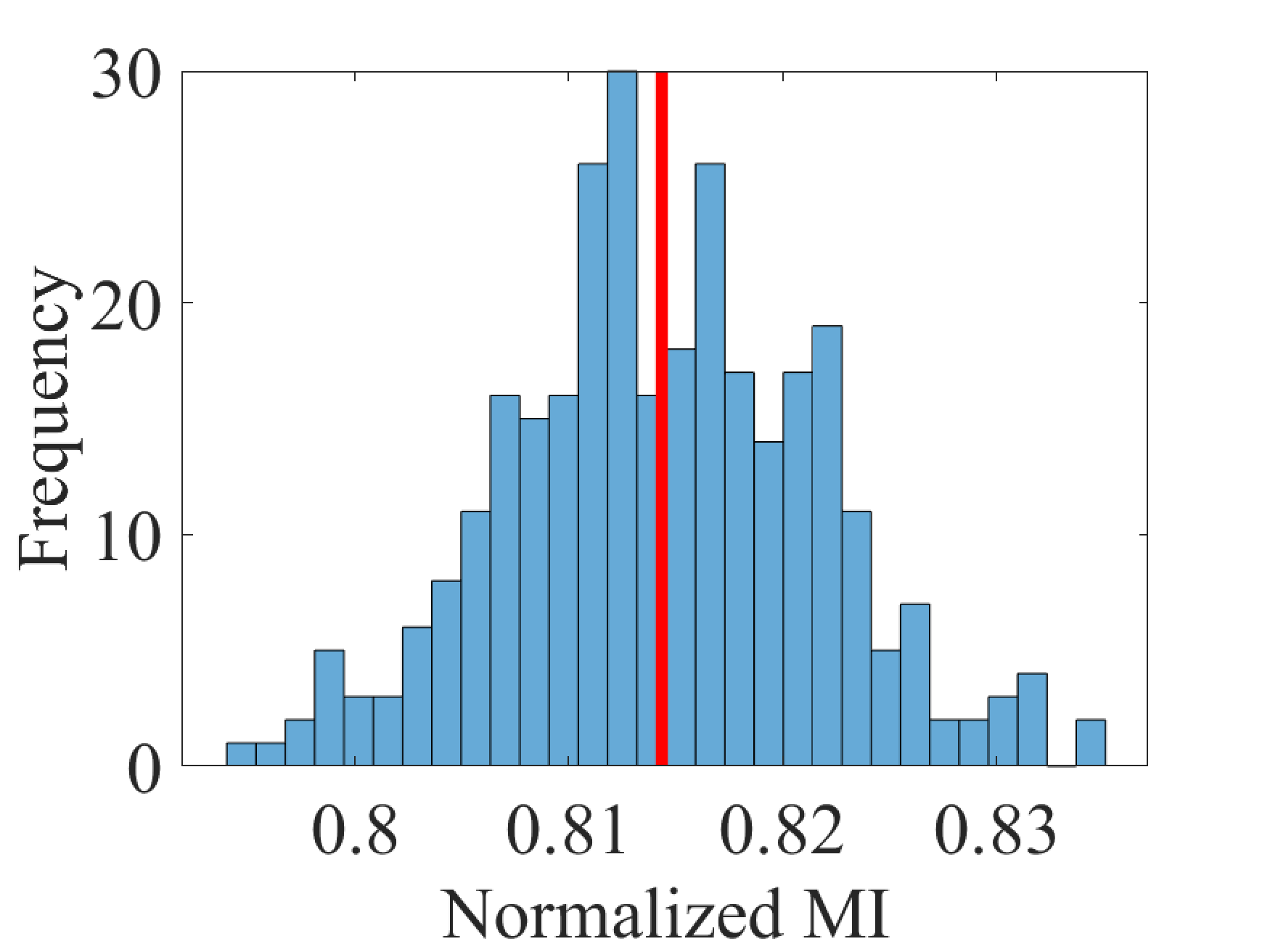}}
	\centerline{(a) \footnotesize \textbf{NST} $n=1$ }\medskip
\end{minipage}
\hspace{0.01mm}
\begin{minipage}[b][2.5cm][t]{0.3\linewidth}
	\centering
	\centerline{\includegraphics[width=3.1cm,height=1.9cm]{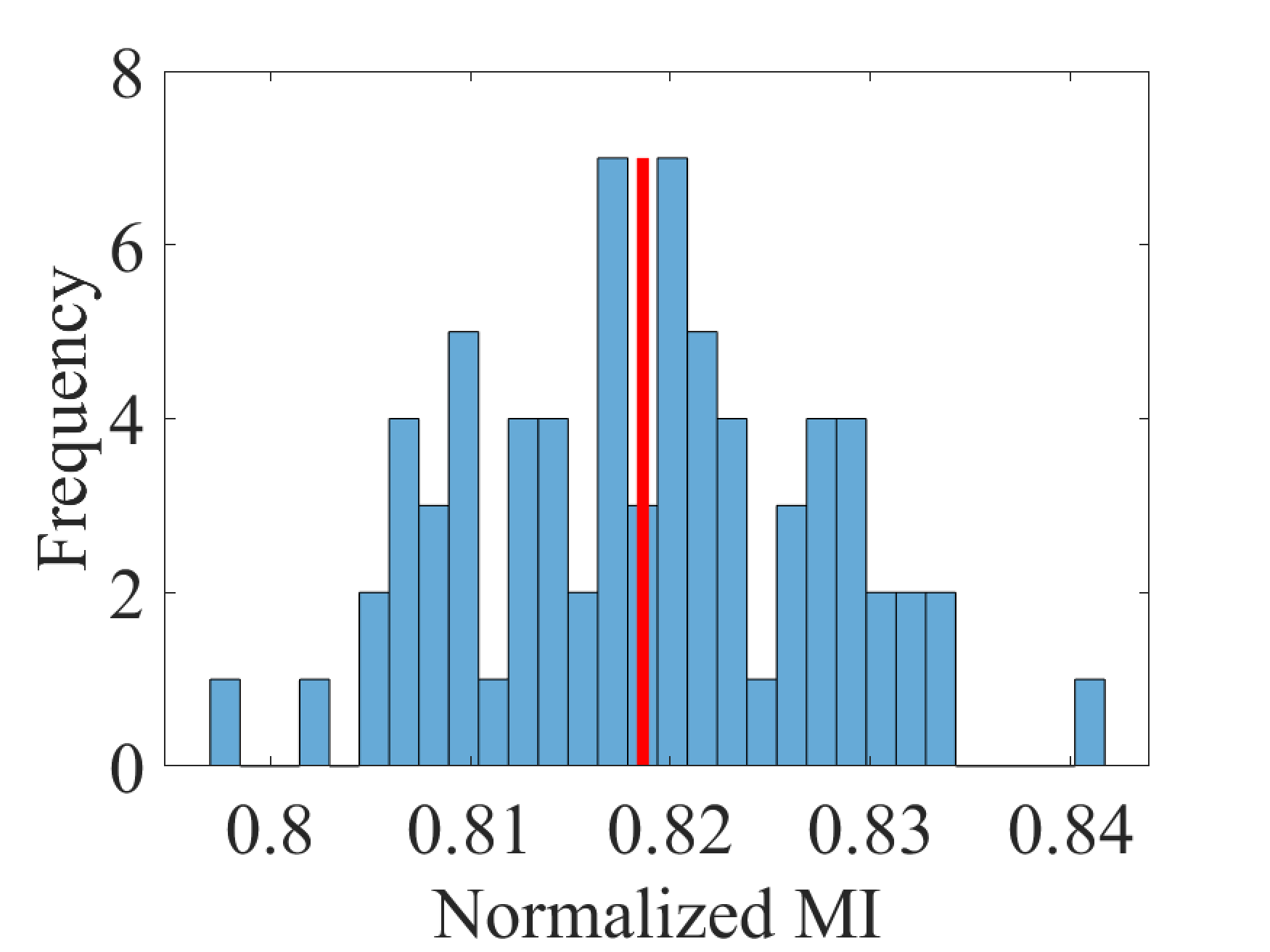}}
	\centerline{(b) \footnotesize \textbf{NST} $n=2$}\medskip
\end{minipage}
\hspace{0.01mm}
\begin{minipage}[b][2.5cm][t]{0.3\linewidth}
	\centering
	\centerline{\includegraphics[width=3.1cm,height=1.9cm]{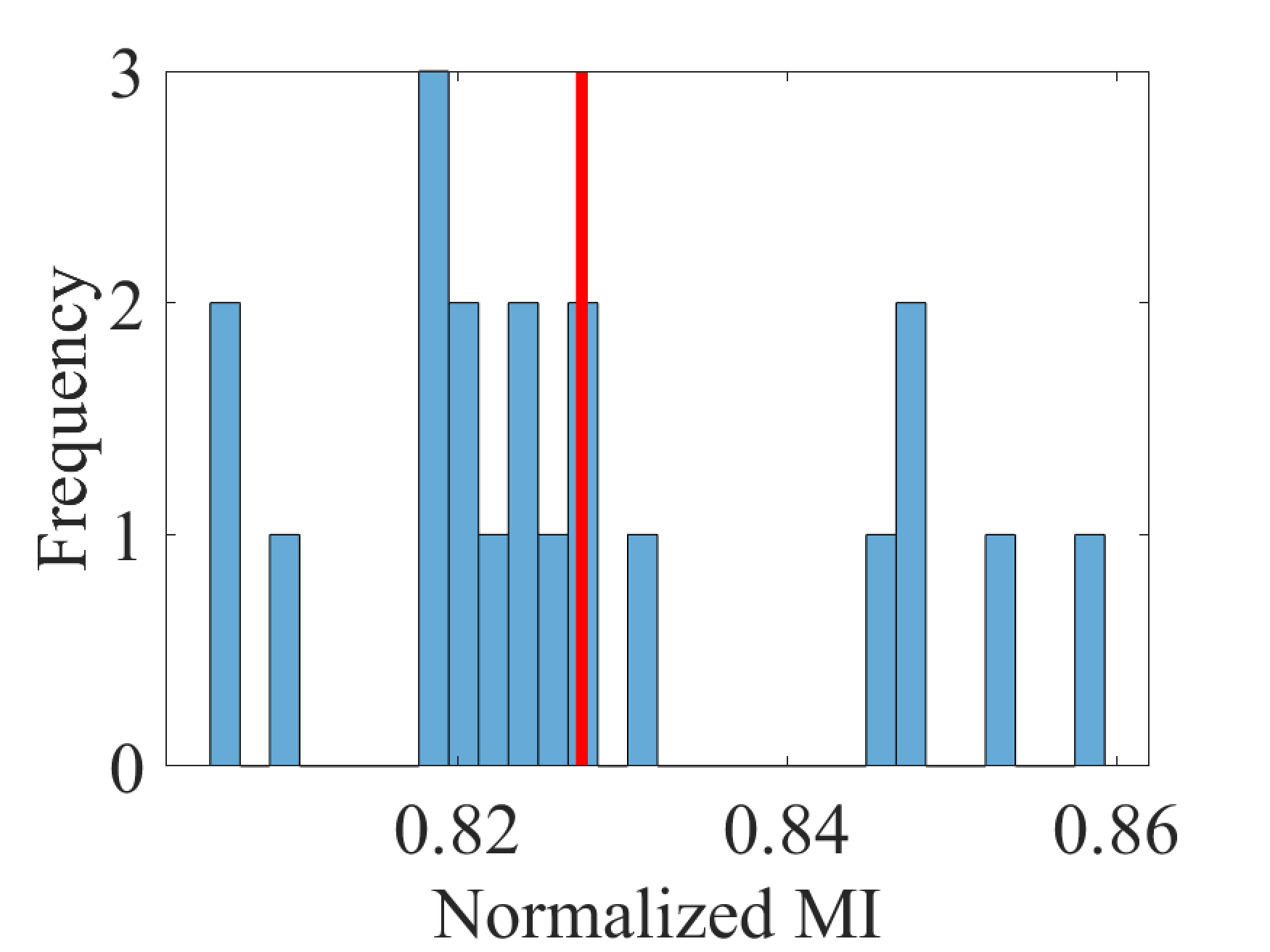}}
	\centerline{(c) \footnotesize \textbf{NST} $n=3$}\medskip
\end{minipage}
\\
\begin{minipage}[b][2.3cm][t]{0.3\linewidth}
	\centering
	\centerline{\includegraphics[width=3.1cm,height=1.9cm]{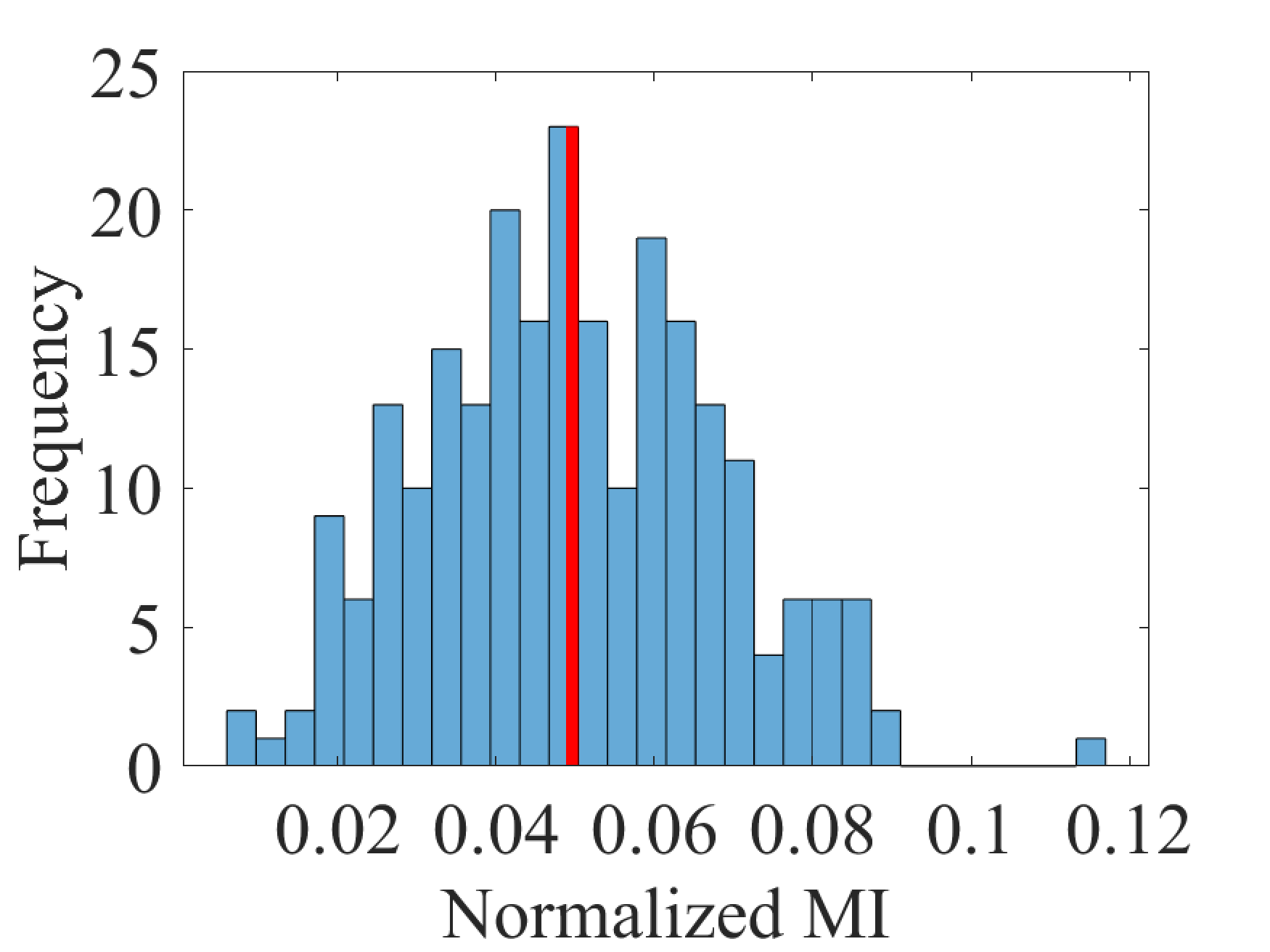}}
	\centerline{(d) \footnotesize \textbf{non-NST} $n=1$}\medskip
\end{minipage}
\hspace{0.01mm}
\begin{minipage}[b][2.3cm][t]{0.3\linewidth}
	\centering
	\centerline{\includegraphics[width=3.1cm,height=1.9cm]{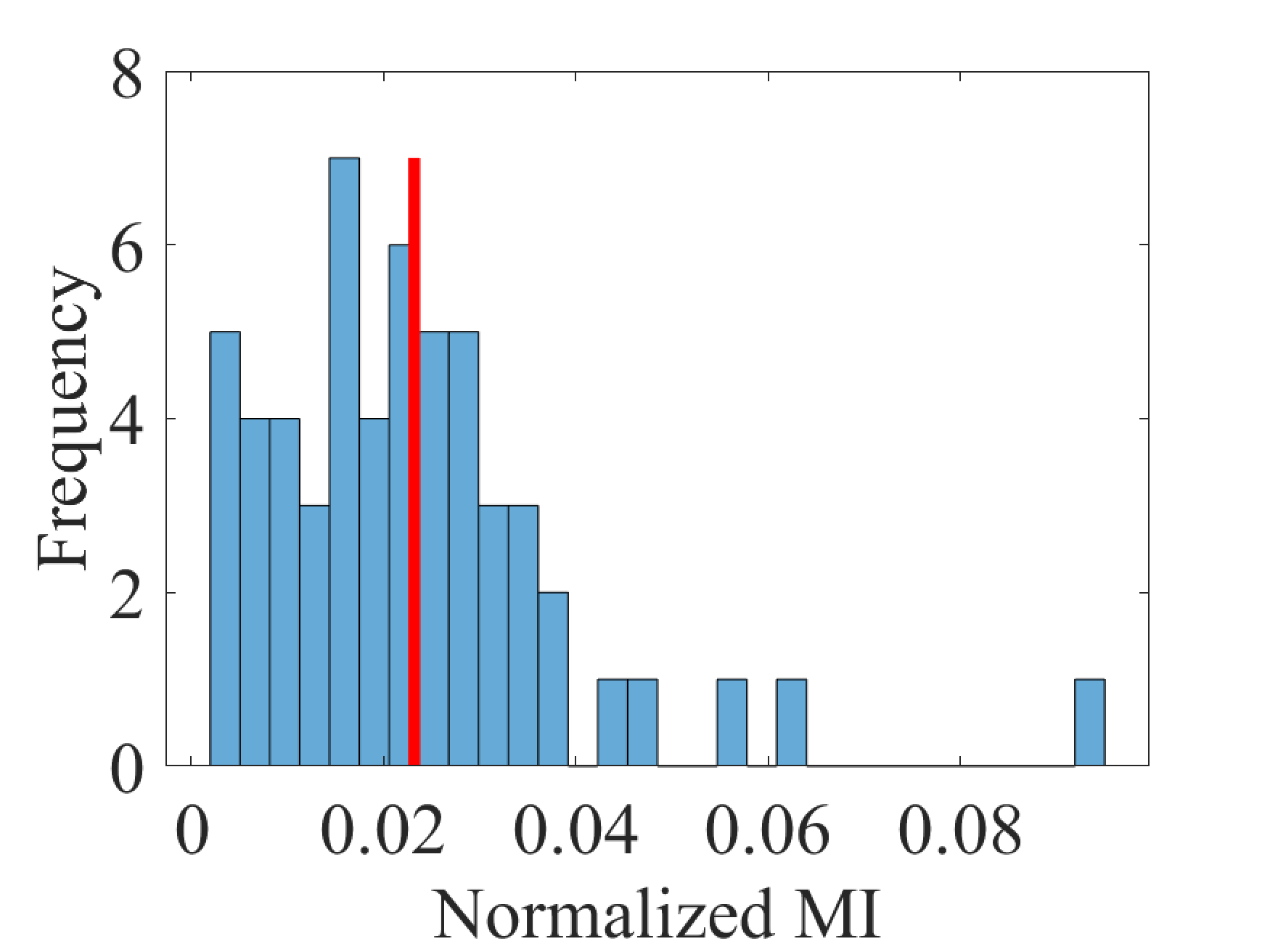}}
	\centerline{(e) \footnotesize \textbf{non-NST} $n=2$}\medskip
\end{minipage}
\hspace{0.01mm}
\begin{minipage}[b][2.3cm][t]{0.3\linewidth}
	\centering
	\centerline{\includegraphics[width=3.1cm,height=1.9cm]{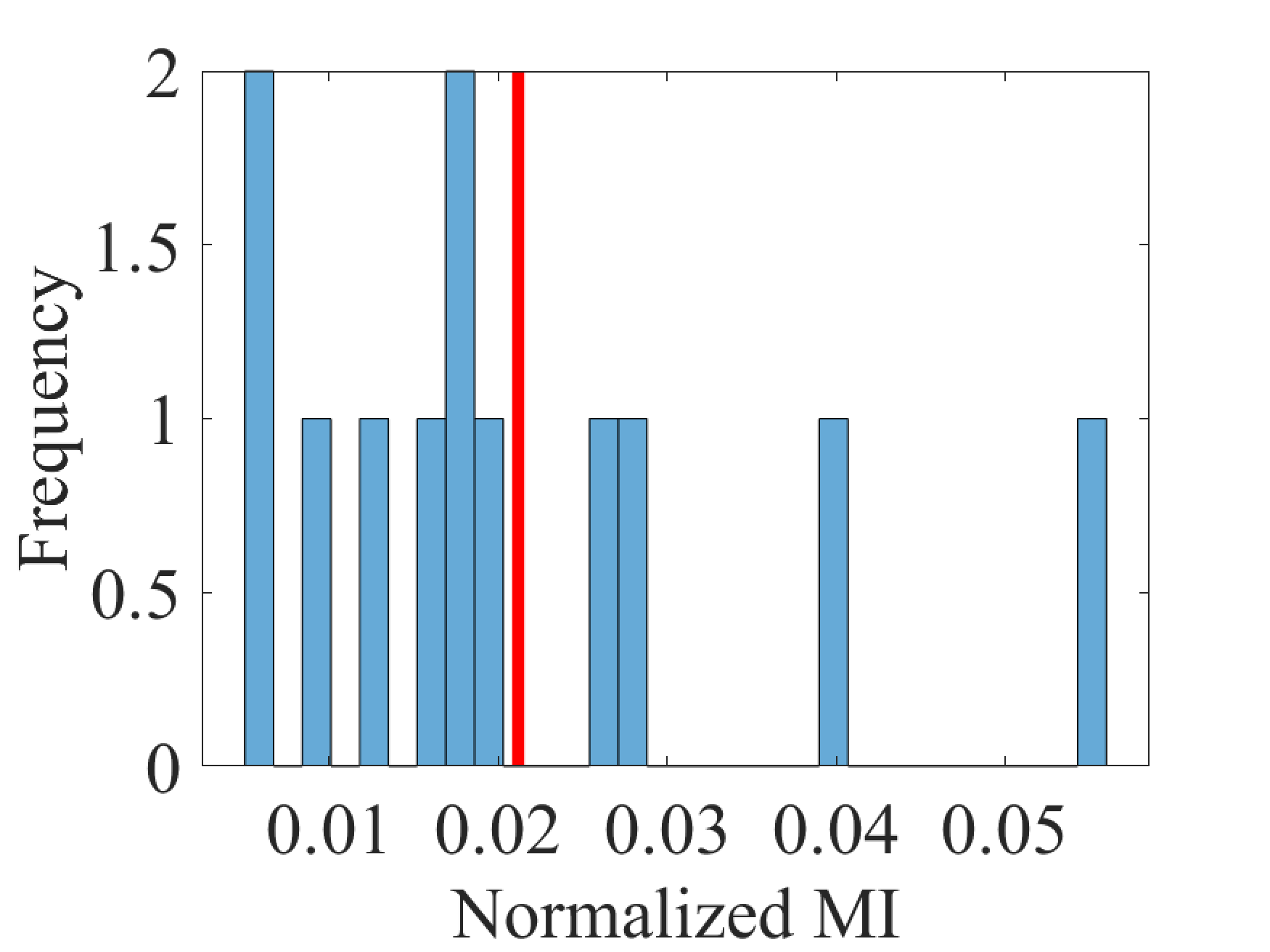}}
	\centerline{(f) \footnotesize \textbf{non-NST} $n=3$ }\medskip
\end{minipage}

\caption{Histograms of MI calculated between patches for 3 different image scales. (a-c) Histograms of MI extracted from pure \textbf{NST} layer for three resolutions $n=1$, $n=2$ and $n=3$. where $n$ refer to the pyramid level. (d-f) Histograms of MI extracted from \textbf{non-NST} layer (the textural data of \textit{camerman}) for three resolutions $n=1$, $n=2$ and $n=3$.}
\label{fig:MIPatches}
\end{figure}

\indent Mutual Information (MI) is used to asses both inter and intra-image similarity \cite{b14,b15}. Here we apply the MI to quantify the self-similarity of texture images. We present two variants of MI-based approaches and evaluate them on non-NST and NST images. The first emphasizes the similarity across \textit{resolutions}, i.e the affinity across scales, which is the essence of Mandelbrot's fractality. The second assesses the self-similarity by quantifying the \textit{spatial} similarity between blocks or patches.\\
\indent The \textit{scale-wise} approach assesses the self-similarity as follows: We first build a pyramidal representation of the textural layer \cite{b13}. We next calculate the MI between two consecutive pyramid levels:
\begin{equation}
MI(I_{n},I_{n-1})=H(I_{n})-H(I_{n}/I_{n-1}),\label{eq:5}
\end{equation}
where $H(I_{n})$ is the entropy of n-th  resolution and $H(I_{n}/I_{n-1})$ is its conditional entropy when the lower resolution counterpart is given. MI measures the amount of information common to two resolutions and yields a degree of similarity between them; High MI values (close enough to the entropy) indicate a high degree of resemblance and vice versa.\\
\indent Comparison between MI of NST and non-NST are highlighted in the table presented in Fig. \ref{fig:MIscales}, where the non-NST layer is residual element extracted from the \textit{cameraman} image (calculated by subtracting the output image of applying PM diffusion from the original image, as described in section \ref{sec:sep}). One can clearly observe  that a pure NST image exhibits high similarity between consecutive pyramid levels, similarly to the referenced fBm image \footnote[1]{We use a synthetic image of fBm field with characteristic $H=0.2$ as a reference of self-similar texture. The simulation of the synthetic fBm images was performed by the FracLab synthesis tool \cite{fraclab}. }, whereas the textural data of cameraman has smaller scale-wise MI values. To observe this property better, we repeat the \textit{scale-wise} calculation of MI for the entire image dataset. The average, minimal and maximal MI values along scales are highlighted in Fig.\ref{fig:MIscales}(b) for the entire set of 240 examples. As one may infer, MI may be instrumental in additional feature in classification of textures.\\  
\indent Pure NST image is considered to be a finite approximation of fBm field, and its self-similarity is defined between scales as expressed in \eqref{eq:4} in a statistical sense. Meaning that the process and its zoomed version share the same statistics. FBm self-similarity expression, using affinity between its blocks or patches, doesn't have closed form, due to its non-stationarity in the spatial domain. However, the \textit{spatial} similarity between the NST sub-blocks can't be neglected and may be characterized by the proposed measure. To this end, we employ the MI measure to quantify the affinity between non-overlapping patches as follows:
 \begin{equation}
MI(P_{i},P_{j})=H(P_{i})-H(P_{i}/P_{l}),
 \end{equation}
where $P_{i}$ and $P_{j}$ are image blocks. (Note that $MI(P_{i},P_{j})=H(P_{i})$ when $i=j$). The \textit{Patch-wise} calculation of MI was used in \cite{b15} to characterize self-similarity of natural images, and its application in fractal-coding was proposed. However, in our framework we use it in a different context, to asses the self-similarity of NST.\\
\indent To investigate the sub-blocks similarity and scale-variance, we performed the patch-wise calculation of MI along 3 image resolutions, obtained by calculating the image pyramid along 3 levels. Pyramid levels, starting from the original image size,  are obtained by reducing the size by 2 at each level, where we divide the images to non-overlapping patches of size $32\times32$. We then calculate the MI normalized by entropy, $MI(P_{i},P_{j})/H(P_{i})$. Histograms of normalized MI for $i\neq j$ are presented in Fig.\ref{fig:MIPatches} for NST and non-NST along three pyramid levels. In contrast to non-NST (texture layer \textit{cameraman}), pure NST shows high MI between its subblocks, which indicates a high  degree of similarity. Preserving this phenomenon along the pyramid levels, implies invariance under resolution. It is worth mentioning that blocks should be sufficiently large to allow reliable statistics estimation, but not too large to preserve spatial local information. In our experiments we use blocks of size $32\times32$.

\subsection{GLCM-based Approach}
\indent In this subsection we proceed to propose an alternative self-similarity measure extracted from the gray level co-occurrence matrix (GLCM) of the textural layer. The GLCM expresses the spatial relationship of image gray levels \cite{b8}. Consider an image with $p$ gray pixel values and of size $m\times n$. The $p\times p$ co-occurrence matrix, $C$, is obtained by:
\begin{equation}
C\left(i,j\right)=\mathop{\sum_{x,y}}G\left(x,y\right),\label{eq:7}
\end{equation}
where 
\begin{equation}
G(x,y)=\begin{cases}
1 &  \text{if}\,I(x,y)=i\:\text{,}\:I(x+d_{x},y+d_{y})=j \\
0       & \text{otherwise}
\end{cases}\label{eq:8}
\end{equation}
$I(x,y)$ is the pixel value at position $(x,y)$ and $(d_{x},d_{y})$ defines the offset, which determine the spatial relation for the GLCM. Having the GLCM, we obtain the probability matrix by normalizing it, $P(i,j)=C(i,j)/\sum_{i,j}C(i,j)$. The probability matrix outlines how a gray level distribution is likely to appear in a specific spatial pattern, predetermined by the offset vector.

\begin{figure}[!b]
	\begin{minipage}[b][4cm][t]{0.45\linewidth}
		\centering
		\centerline{\includegraphics[width=4cm,height=4cm]{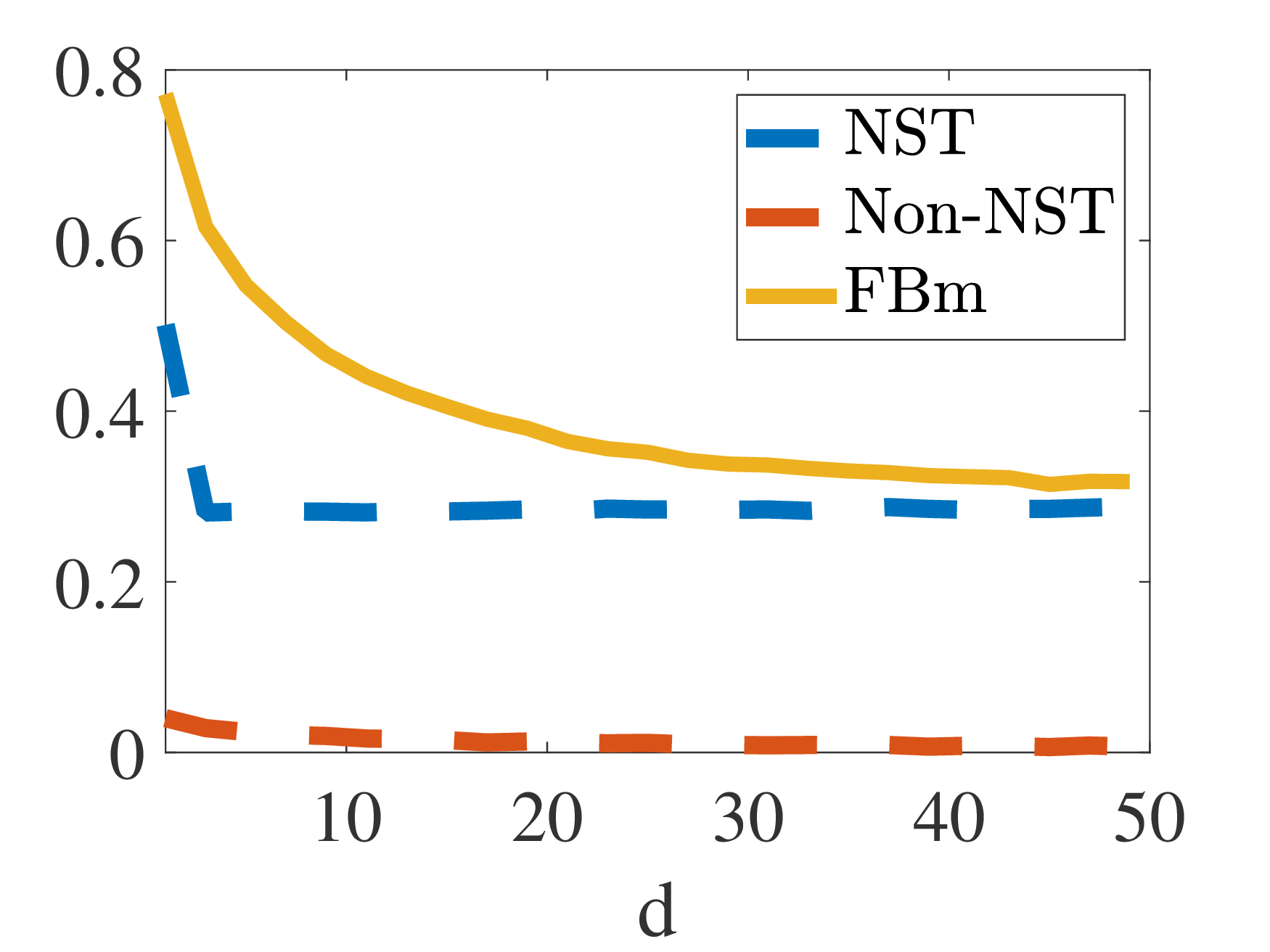}}
	\end{minipage}
	\begin{minipage}[b][4cm][c]{0.45\linewidth}
		\centering
		\scalebox{0.83}[1.3]{
			{\footnotesize{}}%
			\begin{tabular*}{5.3cm}{@{\extracolsep{\fill}}|c|c|c|c|}
				\hline 
				\textbf{\footnotesize{}$\boldsymbol{d_{x},d_{y}}$} & \textbf{\footnotesize{}$2,2$} & \textbf{\footnotesize{}$5,5$} & \textbf{\footnotesize{}$10,10$}\tabularnewline
				\hline 
				\hline 
				\textbf{\footnotesize{}NST} & {\footnotesize{}$0.2852$} & {\footnotesize{}$0.284$} & {\footnotesize{}$0.2836$}\tabularnewline
				\hline 
				\textbf{\footnotesize{}fBm} & {\footnotesize{}$0.6249$} & {\footnotesize{}$0.4993$} & {\footnotesize{}$0.4090$}\tabularnewline
				\hline 
				\textbf{\footnotesize{}non-NST} & {\footnotesize{}$0.0132$} & {\footnotesize{}$0.013$} & {\footnotesize{}$0.0127$}\tabularnewline
				\hline
			\end{tabular*}{\footnotesize\par}}
		\medskip \\
	\end{minipage}
	\caption{GLCM-based Approach evaluation results. Left: MI extracted from GLCM for pure NST (dashed-blue), non-NST (dashed-orange) and synthetic fBm image (solid-yellow), plotted as a function of the offset $d_{x}=d$. Right: MI assessment for thee different values of offset $d_{x}=d_{y}$, pure-NST and fBm images show high MI compared with non-NST, for which we get almost zero MI and its value decrease with increasing pixels distance.}
	\label{fig:MIGLCM}
\end{figure}
 Relevant features (such as contrast, entropy, energy) are extracted from the GLCM to characterize textures\cite{b8,b17}. GLCM-based features are also applied in textures classification \cite{b16}. We extract only the MI term and use it in assessment of self-similarity. MI can be calculated from GLCM as follows:
\begin{equation} 
MI=\sum_{i,j}P(i,j)\log\left(\frac{P(j,j)}{P_{i}(i)P_{i}(j)}\right),
\label{eq:9}
\end{equation}
where $P_{i}(i)=\sum_{j}P(i,j)$ and $P_{j}(j)=\sum_{i}P(i,j)$. This term provides a measure of the information shared between the gray-levels. It quantifies how much knowing one of the variables reduces the uncertainty about knowing the other.\\
\indent How is the MI term expressed in \eqref{eq:9}, related to self-similarity? Intuitively, this statistical term can represent the "impurity" of the image. For complex images, having repeated patterns or textural data, this measure should be high, whereas for images with less periodicity, one might guess that their corresponding MI is smaller.\\
\indent In our experiment we, again, consider the same NST and non-NST examples (textural layers of the image presented in Fig.\ref{fig:Sep} and \textit{cameraman}). The synthetic image of fBm is also used as a reference of self-similar for comparison. We extract the GLCM with various values of offset along $x$ axis only i.e. $d_{y}=0$ and $d_{x}=d$ is a varying parameter. We then calculate the MI, expressed in \eqref{eq:9}. Results of the calculation are summarized in Fig.\ref{fig:MIGLCM}. As one may expect, similarly to the previously discussed quantifiers, this measure also indicates high degree of similarity for NST and its fBm model. Repeating the same  calculation for offsets along the diagonal axis, i.e. $d_{x}=d_{y}=d$, for three values of $d$ are also attached in Fig.\ref{fig:MIGLCM}. The highlighted results imply that the same statistics of self-similarity are not sensitive to the offset selection and this measure could quantify the inherent degree of similarity of the textural data independently of the spatial offset or its direction.     

\section{Discussion and Conclusions}
\indent Natural fully textured images are comprised of both NST and structured information. To asses the properties of Gaussianity, self-similarity and spatial random phase, characteristic of NST, it is therefore necessary to separate first the structured (edge/phase-related) component, which dilutes and contaminate the stochastic nature of texture. To asses the self-similarity we utilized the Mutual information (MI) and expressed the scale invariance by measuring the \textit{scale-wise} similarity. We also extracted the \textit{spatial} affinity of image sub-blocks. Comparison between NST and non-NST cases was provided. Gray-level co-occurrence was used to obtain a MI term that can help in self-similarity quantification.\\
\indent The two approaches reveal the self-similarity of NST and can be therefore used as methods for recognition of NST and non-NST images. The first approach may be preferable due to its simplicity. The GLCM-based approach may be more complex when the gray level precision is high, this may lead to larger GLCM matrix and numerical problems.   

\section*{Acknowledgment}
This research is supported by the Ollendorff Minerva Center. S.K is supported by the Ministry of Science and Technology fellowship.

\end{document}